\documentclass[11pt]{article}

\usepackage[preprint]{acl}
\usepackage{times}
\usepackage{latexsym}
\usepackage[T1]{fontenc}
\usepackage[utf8]{inputenc}
\usepackage{microtype}
\usepackage{inconsolata}
\usepackage{hyperref}
\usepackage{url}
\usepackage{booktabs}
\usepackage{amsfonts}
\usepackage{nicefrac}
\usepackage{xcolor}
\usepackage{amsmath}
\usepackage{enumitem}
\usepackage{multirow}
\usepackage{cleveref}
\usepackage{graphicx}
\usepackage{algorithm}
\usepackage{algorithmic}

\newcommand{\model}{TIAR}

\title{TIAR: Trajectory-Informed Advantage Reweighting for LLM \\ Abstention Learning}

\author{
  Muyu Pan \and Shu Zhao \and Nan Zhang \and Philip Shin \\
  \textbf{Varun Parekh \and Vijaykrishnan Narayanan \and Rui Zhang} \\
  Department of Computer Science, The Pennsylvania State University \\
  State College, PA, USA \\
  \texttt{\{mfp5696, smz5505, njz5124, pws5345, vdp5074, vxn9, rmz5227\}@psu.edu}
}

\begin{document}
\maketitle

\begin{abstract}
  This paper investigates large language model (LLM) abstention learning, specifically using ternary reward, which incentivize truthfulness in large language models. This paper extends that idea by moving from a ternary reward to a Trajectory-Informed advantage reweighting, dynamically re-weights the abstention reward during Group Relative Policy Optimization (GRPO) training. The objective of this work focuses on abstention learning instead of improving truthfulness, serving as an exploration into hallucination reduction. The novelty of this paper lies in methodological innovation, advantage re-weighting, and benchmark selection. Leveraging GRPO's multiple trajectories as a natural abstention signal, this method uses a reward signal to explore knowledge boundaries and encourage consistency. By demonstrating that trajectories can be used as a confidence indicator of the policy relative to the query, they are then used to dynamically calculate the abstention advantage. AbstentionBench is used as the evaluation benchmark, as this work aims to contribute to the field of abstention learning. All datasets on the benchmark were tested against this method and various baselines. Empirical results demonstrate that TIAR achieves state-of-the-art abstention F1 scores across five of six evaluation categories, outperforming the static ternary baseline on 17 of 31 benchmark datasets while fully preserving baseline accuracy.
\end{abstract}
\footnotetext{Our code and resources are available at: \url{https://anonymous.4open.science/r/TIAR_repo/}.}

\section{Introduction}

Abstention learning has become a main theme in assisting hallucination reduction in large language models. The purpose of abstention learning is to teach LLMs to say "I don't know" for questions that they do not have the ability to answer. Such behavior benefits high-stakes domain applications, such as in medicine, when a query relates to the personal safety of a patient and is difficult to answer, in this case, abstention is preferred over hallucination for safety considerations. In the recent work "Why Large Language Models Hallucinate" by OpenAI  \citep{kalai2025languagemodelshallucinate}, the authors mention the importance of using abstention to mitigate hallucination and call for further exploration of abstention learning. Previous work \citep{wen2025knowlimitssurveyabstention} summarizes the current progress in abstention learning, spanning alignment and inference.

In alignment, supervised fine-tuning methods such as R-Tuning \citep{zhang2024rtuninginstructinglargelanguage} prepare datasets where labels for unanswerable questions are rewritten as "I don't know." While fine-tuning reduces hallucinations in specific reasoning tasks, such as logical translation  \citep{lang2logic}, abstention offers a generalized safety mechanism when models reach their broad knowledge boundaries. Preference learning exploration  \citep{rafailov2024directpreferenceoptimizationlanguage} similarly prepares offline reinforcement learning datasets with sets of answers that include an "I don't know" option. Another category focuses on inference-stage abstention, such as sampling the same query multiple times to estimate confidence and decide whether to abstain based on the result  \citep{cole2023selectivelyansweringambiguousquestions, lin2022teachingmodelsexpressuncertainty, phute2024llmselfdefenseself, zhao2022calibratingsequencelikelihoodimproves, slobodkin2023curiouscasehallucinatoryunanswerability}.

The state-of-the-art method in the alignment approach is TruthRL \citep{wei2025truthrlincentivizingtruthfulllms}, which uses GRPO training with a ternary reward (correct trajectory: $+1$, abstention trajectory: $0$, incorrect trajectory: $-1$) to incentivize model truthfulness, representing hallucination reduction with improved accuracy. Previous work \citep{kadavath2022languagemodelsmostlyknow} showed that question difficulty affects a model's response accuracy, and during GRPO training, the model tends to converge to a single trajectory that provides the highest advantage for a given query, indicating increasingly consistent outputs through training \citep{shao2024deepseekmathpushinglimitsmathematical}. To better teach abstention, we use GRPO's multiple sampled trajectories serve as a natural signal for abstention: for consistent trajectories, the decision is straightforward. For mixed trajectories, a reward signal can be used to explore the knowledge boundary and encourage consistency.

AbstentionBench \citep{kirichenko2025abstentionbenchreasoningllmsfail} is a comprehensive benchmark that tests abstention ability across 20 datasets and 6 scenarios. It uses abstention F1, recall, and precision as main metrics, and provides accuracy as a complementary metric. To demonstrate generalizability, empirical evaluation is performed on AbstentionBench across all datasets. Abstention F1 reflects accurate abstention while avoiding over-abstention, and accuracy measures correctness on questions with reference answers. Since AbstentionBench showed no direct relationship between abstention F1 and accuracy, our goal is to improve abstention while preserving accuracy as an independent capability indicator. Compared to baselines, TIAR achieves better abstention F1 and accuracy on most datasets, indicating it is a more effective abstention learning method that improves the model on both fronts. 

Overall, our proposed method TIAR is a simple and practical approach that harnesses the natural advantage of GRPO to dynamically estimate self-confidence and teach LLM abstention, rather than relying on a static reward function when query difficulty varies. Empirically, TIAR achieves state-of-the-art abstention F1 scores across five of six evaluation categories and outperforms the static ternary baseline on 17 of 31 AbstentionBench datasets, all while preserving crucial baseline accuracy. Grounded in rigorous derivation, we believe this quantifiable improvement represents a significant step forward in the direction of abstention learning, providing valuable insights for future work.

\section{Related Works}
\label{gen_inst}

\textbf{Abstention Learning} Abstention learning is defined as teaching LLMs to refuse unanswerable questions, gradually becoming a key field for hallucination mitigation. Supervised Fine-Tuning approaches like Alignment for Honesty \citep{yang2024alignmenthonesty} and R-tuning prepare datasets explicitly labeling answerable questions as "I don't know". Offline Reinforcement Learning approaches \citep{cheng2024aiassistantsknowdont} prepare datasets containing sets of answers to each query, learning to pick the "I don't know" choice through preference learning. GRPO approaches are among the most effective. One approach \citep{gul2025paypersearchmodelsabstentionmodels} first fine-tunes the model to use searching tools for unanswerable questions, later replacing them with abstention during GRPO. Other approaches innovate on the reward function, showing that traditional binary rewards encourage guessing, while ternary rewards successfully reduce hallucination rates by 19.1\% and improve truthfulness  \citep{wei2025truthrlincentivizingtruthfulllms}. Alternative strategies include decoding-based  \citep{Chae_2024}, Conformal Prediction  \citep{tayebati2025learningconformalabstentionpolicies}, and Uncertainty Estimation methods  \citep{cohen2024idontknowexplicit}. TIAR uniquely uses GRPO trajectories as a natural abstention signal to explore knowledge boundaries, dynamically reweighting advantages based on query difficulty. Our objective focuses on abstention learning, rather than directly improving truthfulness.

\textbf{Reinforcement learning for Hallucination Mitigation} Reinforcement Learning is widely applied for LLM hallucination mitigation, traditional pre-training objective of maximizing likelihood focuses on the next token prediction but not factual accuracy, it tends to generate next token that calibrated from the learned data distribution from pre-training. \citep{ouyang2022traininglanguagemodelsfollow} Reinforcement Learning from Human Feedback (RLHF) incentivizes this behavior by training the model to calibrate with human preferred response style, which make the model's response confident and sound but at the expense of factual accuracy. \citep{openai2024gpt4technicalreport} Reinforcement Learning with Verifiable Rewards (RLVR) paradigms using binary reward signals exacerbate this behavior to train the model to be a good test taker, encouraging guessing over abstention. Beyond hallucination mitigation, RL and fine-tuning signals are increasingly leveraged to improve other complex capabilities of LLMs, ranging from the quantization of large reasoning models  \citep{quantlrm} to sophisticated NLP and RL frameworks applied to SAT solving  \citep{langsat}. 

Beyond the abstention-specific methods, under the behavioral calibration approach, previous work  \citep{an2025teachingllmsabstainfinegrained} applies Semantic Clustering to GRPO trajectories, determining correct trajectories by comparing cluster sizes against a manually set threshold. TIAR differs by avoiding explicit Semantic Clustering and manual thresholds, offering a simpler reward design that effectively boosts both abstention F1 and accuracy. Another work  \citep{wu2026mitigatingllmhallucinationbehaviorally} requires users to explicitly specify a risk score $t \in [0, 1]$, scaling the ternary abstention reward to $2t - 1$, where high-risk queries encourage abstention. TIAR differs by eliminating the need for manual risk inputs, dynamically adapting to query difficulty on its own, and focusing strictly on abstention learning rather than broad hallucination mitigation under varying risks.

\section{Methodology}
 
\subsection{Preliminaries}
 
Knowledge boundary probing refers to constructing a dataset containing both answerable and unanswerable questions relative to the model's capabilities. For each training question, multiple responses are sampled from the LLM; if none are correct, the question is marked as out-of-knowledge (OOK) and labeled with ``I don't know'' as the ground-truth answer. This approach has been explored in prior works such as R-Tuning~ \citep{zhang2024rtuninginstructinglargelanguage}.

TIAR is implemented using the online RL algorithm GRPO~ \citep{shao2024deepseekmathpushinglimitsmathematical}, which optimizes:

\begin{align}
&\mathcal{L}_{\text{GRPO}}(\theta) = \nonumber\\
&-\mathbb{E}_{x,\{y_i\}_{i=1}^G}\!\left[\frac{1}{G}\sum_{i=1}^{G}\frac{1}{|y_i|}\sum_{t=1}^{|y_i|}
\min\!\Big(w_{i,t}\hat{A}_i,\right.\nonumber\\
&\qquad\left.\text{clip}(w_{i,t}, 1{-}\epsilon, 1{+}\epsilon)\hat{A}_i\Big)
- \beta D_{\text{KL}}(\pi_\theta \| \pi_{\text{ref}})\right],
\label{eq:grpo}
\end{align}

where $\mathcal{D}$ is the dataset distribution, $\theta$ represents the model parameters, $x \sim \mathcal{D}$, $\{y_i\}_{i=1}^G \sim \pi_{\theta_{\text{old}}}(\cdot|x)$, $\epsilon$ and $\beta$ are hyperparameters, $G$ is the group size (number of sampled responses per query), $\pi_{\text{ref}}$ is the reference policy, $w_{i,t} = w_{i,t}(\theta)$ is the importance ratio, $D_{\text{KL}}$ is the Kullback-Leibler divergence, and $\hat{A}_i$ is the estimated advantage for response $y_i$, computed via group-level normalization:

\begin{equation}
\hat{A}_i = \frac{r(x, y_i) - \text{mean}\{r(x, y_j)\}_{j=1}^{G}}{\text{std}\{r(x, y_j)\}_{j=1}^{G}}.
\label{eq:grpo_advantage}
\end{equation}
 
\subsection{Reward Structure and the Abstention Decision}
 
Consider a model facing a question $x$ with group size $G$. Among the $G$ sampled trajectories, let $n_c$, $n_w$, and $n_a$ denote the number of correct, wrong, and abstention trajectories respectively, with $n_c + n_w + n_a = G$. Under the ternary reward scheme of TruthRL~ \citep{wei2025truthrlincentivizingtruthfulllms}:

\begin{equation}
R_c = +1, \quad R_w = -1, \quad R_a = 0.
\label{eq:ternary}
\end{equation}
 
We define the model's empirical correctness rate among attempted (non-abstaining) trajectories as:

\begin{equation}
\hat{p} = \frac{n_c}{n_c + n_w}, \quad \text{when } n_c + n_w > 0.
\label{eq:phat}
\end{equation}

This quantity estimates the model's probability of answering correctly \emph{given that it attempts}, and serves as a natural proxy for question difficulty relative to the current policy: high $\hat{p}$ indicates an easy question, while low $\hat{p}$ indicates a hard one.

\paragraph{When should the model abstain?}
From the model's perspective, the expected value of attempting to answer is:
\begin{equation}
V_{\text{attempt}}(\hat{p}) = \hat{p} \cdot R_c + (1 - \hat{p}) \cdot R_w = 2\hat{p} - 1.
\label{eq:vattempt}
\end{equation}
The model should prefer abstention (receiving $R_a$) over attempting when $R_a > V_{\text{attempt}}(\hat{p})$. Under the ternary reward ($R_a = 0$), this becomes $0 > 2\hat{p} - 1$, i.e., $\hat{p} < 0.5$. The ternary reward thus implicitly sets a fixed abstention threshold at $\hat{p}^* = 0.5$, regardless of the actual difficulty distribution encountered during training.

\paragraph{The opportunity cost of abstention.}
We derive the optimal abstention reward by computing the net value of abstaining versus attempting. When the model abstains, it avoids the expected loss $(1{-}\hat{p})\cdot|R_w|$ but omits the expected gain $\hat{p} \cdot R_c$. The net value is:
\begin{equation}
R_a^* = \underbrace{(1 - \hat{p}) \cdot |R_w|}_{\text{loss avoided}} - \underbrace{\hat{p} \cdot R_c}_{\text{gain\ costed}} = 1 - 2\hat{p}.
\label{eq:optimal_ra}
\end{equation}
 
This dynamic reward has the following properties:
\begin{itemize}[nosep]
    \item $\hat{p} = 0$ (all attempts fail): $R_a^* = 1$, abstention is maximally rewarded.
    \item $\hat{p} = 0.5$: $R_a^* = 0$, recovering the ternary reward as a special case.
    \item $\hat{p} = 1$ (all attempts succeed): $R_a^* = -1$, abstention is maximally penalized.
\end{itemize}
 
\noindent The ternary reward is thus a special case of the dynamic reward when $\hat{p} = 0.5$, treating all questions as equally difficult, an assumption that does not hold in practice.

\subsection{The Coupling Problem in GRPO}
\label{sec:coupling}
 
A natural approach would be to directly substitute $R_a = 1 - 2\hat{p}$ into the reward and compute GRPO advantages via Eq.~\ref{eq:grpo_advantage}. However, this introduces a \emph{coupling problem}: because all advantages share the same group mean $\bar{R}$ and standard deviation $\sigma$, modifying $R_a$ changes the normalization statistics, which distorts $\hat{A}_c$ and $\hat{A}_w$. The group mean under ternary rewards is:
\begin{equation}
\bar{R}_{\text{ternary}} = \frac{n_c - n_w}{G},
\end{equation}
while under the dynamic reward $R_a = 1 - 2\hat{p}$:
\begin{equation}
\bar{R}_{\text{dynamic}} = \bar{R}_{\text{ternary}} + \frac{n_a(1 - 2\hat{p})}{G}.
\end{equation}
 
On hard questions ($\hat{p} < 0.5$), the dynamic reward is positive, pushing $\bar{R}_{\text{dynamic}} > \bar{R}_{\text{ternary}}$. This reduces the advantage for correct trajectories:
\begin{equation}
\hat{A}_c^{\text{dynamic}} = \frac{R_c - \bar{R}_{\text{dynamic}}}{\sigma_{\text{dynamic}}} < \frac{R_c - \bar{R}_{\text{ternary}}}{\sigma_{\text{ternary}}} = \hat{A}_c^{\text{ternary}}.
\end{equation}
where $\sigma_{\text{dynamic}}$ and $\sigma_{\text{ternary}}$ represent the standard deviations of the respective reward schemes. This is undesirable: on hard questions where correct answers are rare and valuable, the coupled approach weakens the learning signal for those correct trajectories. We empirically confirm this: a coupled implementation degrades accuracy by $-0.76\%$ compared to the ternary baseline, despite improving abstention F1 by $+0.15\%$.

\begin{table}[h]
\caption{Coupled reward modification ($R_a = 1 - 2\hat{p}$) vs.\ ternary baseline (TruthRL) at step~20, averaged across all AbstentionBench datasets.}
\label{tab:coupling_empirical}
\vskip 0.1in
\centering
\small
\resizebox{\columnwidth}{!}{
\begin{tabular}{lcc}
\toprule
Method & Acc. & Abstention F1 \\
\midrule
TruthRL (ternary) & 67.07 & 72.45 \\
Coupled & 66.31 & 72.60 \\
\midrule
$\Delta$ & $-$0.76 & +0.15 \\
Win / Loss & 6/16 & 13/13 \\
\bottomrule
\end{tabular}
}
\vskip -0.1in
\end{table}
 
\subsection{TIAR: Decoupled Advantage Adjustment}
\label{sec:TIAR}

\begin{figure*}[t]
  \centering
  \includegraphics[width=\textwidth]{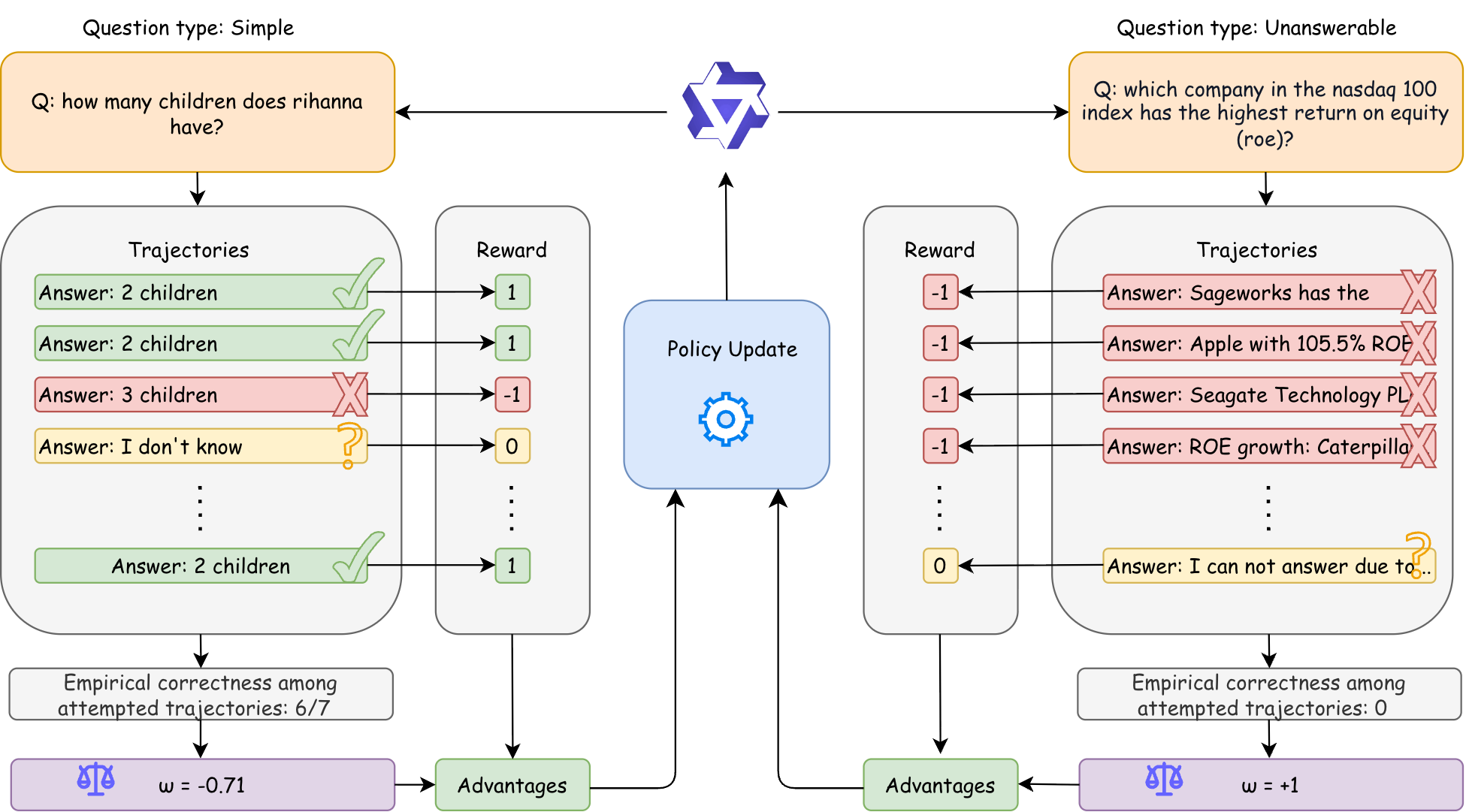}
  \caption{Abstention trajectories are reweighted based on the empirical correctness among attempted trajectories. \model{} discourages abstention for simple questions and boosts the advantages for unanswerable questions.}
  \label{fig:quant}
\end{figure*}
 
To resolve the coupling problem, we propose \textbf{TIAR} (Trajectory-Informed Advantage Reweighting), which applies the derived insight as a post-hoc advantage adjustment rather than a reward modification. The key idea is to preserve the standard GRPO normalization for correct and incorrect trajectories while separately adjusting the abstention advantage based on the group's empirical difficulty.

\paragraph{Algorithm.}
Given a GRPO training step with group size $G$, the specific mechanism is outlined in Algorithm~\ref{alg:TIAR}. The overall framework is illustrated in Figure~\ref{fig:quant}.

\begin{algorithm}[h]
\caption{TIAR: Decoupled Advantage Adjustment}
\label{alg:TIAR}
\begin{algorithmic}[1]
\REQUIRE Query batch $\{x_i\}$, group size $G$, strength $\lambda$
\FOR{each query $x_i$ in batch}
    \STATE Sample $G$ trajectories $\{y_1, \ldots, y_G\} \sim \pi_{\theta_{\text{old}}}(\cdot | x_i)$
    \STATE Assign ternary rewards: $r_j \in \{+1, -1, 0\}$
    \STATE Compute GRPO advantages via Eq.~\ref{eq:grpo_advantage}
    \STATE $n_c \leftarrow |\{j : r_j {=} 1\}|$, $n_w \leftarrow |\{j : r_j {=} {-}1\}|$
    \IF{$n_c + n_w > 0$}
        \STATE $\hat{p} \leftarrow n_c / (n_c + n_w)$
        \FOR{each $j$ where $r_j = 0$}
            \STATE $\hat{A}_j \leftarrow \hat{A}_j + \lambda(1 - 2\hat{p})$
        \ENDFOR
    \ENDIF
\ENDFOR
\STATE Update $\pi_\theta$ with modified advantages (Eq.~\ref{eq:grpo})
\end{algorithmic}
\end{algorithm}

\paragraph{Connection to the opportunity cost derivation.}
The adjustment $\lambda(1 - 2\hat{p})$ is the difference between the optimal and ternary rewards:
\begin{equation}
\Delta_a = R_a^* - R_a^{\text{ternary}} = (1 - 2\hat{p}) - 0 = 1 - 2\hat{p},
\end{equation}
applied as a post-normalization adjustment. The scaling factor $\lambda$ controls the correction strength. In all experiments, we use $\lambda = 1.0$.

\section{Experiment Setup}

\subsection{Training}

\textbf{Base model and dataset.}
We use Llama-3.1-8B-Instruct~ \citep{grattafiori2024llama3herdmodels} and Qwen3-8B \citep{yang2025qwen3technicalreport} as the base models for all methods. For GRPO training, we use the TruthRL-CRAG dataset~ \citep{wei2025truthrlincentivizingtruthfulllms, yang2024cragcomprehensiverag}, which contains 656 training samples derived from the CRAG benchmark. Although all training questions have ground-truth answers, many lie beyond the model's parametric knowledge, creating a natural distribution of answerable and unanswerable questions relative to the model's capabilities.

\textbf{GRPO configuration.}
Training is conducted using the verl framework~ \citep{Sheng_2025} with GRPO as the advantage estimator. We use a batch size of 64 and a group size (rollout $n$) of 8. The maximum prompt length is 16,384 tokens and maximum response length is 2,048 tokens. Hyperparameters such as the learning rate ($1\times10^{-6}$) and the KL loss coefficient ($0.001$, with low-variance KL divergence) were directly adopted from the TruthRL baseline settings without further tuning to ensure a fair comparison. The scaling parameter $\lambda = 1.0$ for TIAR was determined via our ablation study.

We use vLLM~ \citep{kwon2023efficientmemorymanagementlarge} as the rollout engine with tensor parallel size of 2 and GPU memory utilization of 0.8. FSDP is used for actor training with gradient checkpointing enabled. Training runs on 16 NVIDIA A100 40GB GPUs managed via SLURM. Training for 20 steps took approximately 6.7 hours. Total computational budget for a single model is 107.2 GPU-hours. Due to the high computational cost of GRPO, all reported results reflect a single training run with a fixed random seed.

\textbf{Reward function.} An external LLM judge (Llama-3.1-8B-Instruct served via vLLM) evaluates each generated response and assigns ternary rewards: $+1$ for correct, $-1$ for incorrect, and $0$ for abstention. The judge classifies responses as abstentions using pattern matching and semantic analysis, then evaluates correctness of non-abstention responses against ground-truth answers.

\textbf{Baselines.} We compare TIAR against the following methods, all using Llama-3.1-8B-Instruct:
\begin{itemize}
    \item \textbf{R-Tuning}: Supervised fine-tuning on the CRAG dataset with out-of-knowledge questions relabeled as "I don't know" via knowledge boundary probing, sampling 256 responses per question and marking questions as out-of-knowledge if none is correct.
    \item \textbf{RFT} (Rejection Fine-Tuning): Same knowledge boundary probing as SFT, but fine-tunes only on correctly answered responses, filtering out incorrect ones.
    \item \textbf{DPO}~ \citep{rafailov2024directpreferenceoptimizationlanguage}: Trains on preference pairs where ``I don't know'' is preferred over incorrect answers for out-of-knowledge questions.
    \item \textbf{TruthRL}~ \citep{wei2025truthrlincentivizingtruthfulllms}: GRPO with ternary rewards ($R_c{=}1, R_w{=}{-}1, R_a{=}0$) using the same training configuration as TIAR, serving as the direct ablation that isolates the effect of the advantage adjustment.
\end{itemize}

\subsection{Evaluation}

\textbf{Benchmark.}
We evaluate on AbstentionBench~ \citep{kirichenko2025abstentionbenchreasoningllmsfail}, a comprehensive benchmark containing 20 datasets across 31 subsets spanning six abstention scenarios: answer unknown, underspecified intent, stale information, underspecified context, false premise, and subjective questions.

\textbf{Metrics.}
Following AbstentionBench, we report four metrics: \textbf{Abstention F1} (harmonic mean of abstention precision and recall), \textbf{Abstention Recall} (proportion of correctly abstained unanswerable questions), \textbf{Abstention Precision} (proportion of warranted abstentions), and \textbf{Accuracy} (correctness of responses across all samples with reference answers). Abstention F1 is the primary metric reflecting the quality of abstention decisions, while accuracy serves as an independent indicator of the model's general capability.

\textbf{Evaluation protocol.} All models are evaluated using vLLM inference with temperature 0.8, top-$p$ 0.95, and maximum generation length of 4,096 tokens. An LLM-as-Judge pipeline (Llama-3.1-8B-Instruct) performs two-stage evaluation: first classifying each response as abstention or non-abstention, then evaluating factual correctness of non-abstention responses against ground truth. This judge was validated against human annotation at 82.3\% accuracy by the AbstentionBench authors.

\section{Results}

\subsection{Main Results}

\begin{table*}[t!]
\caption{Abstention F1, Recall, Precision, and Accuracy across six AbstentionBench categories (one representative dataset per category). Best scores per metric per model are \textbf{bold}.}
\label{tab:main_results}
\centering
\small
\vspace{6pt}
\resizebox{\textwidth}{!}{
\begin{tabular}{clcccccc}
\toprule
\multicolumn{2}{c}{Method} & \multicolumn{6}{c}{Representative Dataset (AbstentionBench Category)} \\
\cmidrule(lr){1-2}\cmidrule(lr){3-8}
 & & BB/Known Unk. & BBQ & FreshQA & UMWP & QAQA & KUQ/Cont. \\
Model & Metric & {\scriptsize Answer Unknown} & {\scriptsize Underspecified Intent} & {\scriptsize Stale} & {\scriptsize Underspecified Context} & {\scriptsize False Premise} & {\scriptsize Subjective} \\
\midrule
\multicolumn{8}{l}{\textit{Llama-3.1-8B-Instruct}} \\
\midrule
R-Tuning & Abstention F1 & 91.7 & 82.4 & 69.9 & 75.3 & 56.0 & 75.0 \\
 & Abstention Recall & 95.7 & 90.6 & 66.1 & 60.7 & 45.6 & 67.7 \\
 & Abstention Precision & 88.0 & 75.5 & 74.1 & 99.1 & 72.6 & 84.1 \\
 & Accuracy & \textbf{91.3} & 62.7 & 40.0 & 86.2 & 44.2 & 66.3 \\
\midrule
RFT & Abstention F1 & \textbf{97.9} & \textbf{85.9} & 72.2 & 76.1 & 57.1 & \textbf{77.7} \\
 & Abstention Recall & \textbf{100.0} & 84.8 & 66.7 & 61.7 & 45.6 & \textbf{70.9} \\
 & Abstention Precision & \textbf{95.8} & \textbf{87.2} & \textbf{78.7} & \textbf{99.2} & 76.5 & \textbf{86.0} \\
 & Accuracy & 87.0 & 67.6 & 50.0 & 86.7 & 44.9 & 67.8 \\
\midrule
DPO & Abstention F1 & 93.9 & 84.4 & 72.1 & 75.5 & 58.0 & 74.6 \\
 & Abstention Recall & \textbf{100.0} & 91.2 & 70.1 & 61.1 & \textbf{47.0} & 68.2 \\
 & Abstention Precision & 88.5 & 78.5 & 74.3 & 98.7 & 75.7 & 82.1 \\
 & Accuracy & 69.6 & 67.4 & 48.0 & 89.7 & 46.7 & 66.7 \\
\midrule
TruthRL & Abstention F1 & 95.8 & 84.5 & \textbf{74.7} & \textbf{78.9} & 58.0 & 74.8 \\
 & Abstention Recall & \textbf{100.0} & 91.5 & \textbf{75.7} & \textbf{65.5} & 46.7 & 66.5 \\
 & Abstention Precision & 92.0 & 78.5 & 73.6 & \textbf{99.2} & 76.4 & 85.5 \\
 & Accuracy & 87.0 & \textbf{71.7} & 54.0 & 90.3 & \textbf{50.5} & \textbf{70.7} \\
\midrule
TIAR (Ours) & Abstention F1 & \textbf{97.9} & 84.6 & 74.5 & 77.6 & \textbf{58.2} & 76.2 \\
 & Abstention Recall & \textbf{100.0} & \textbf{92.4} & 75.1 & 63.8 & 46.7 & 69.7 \\
 & Abstention Precision & \textbf{95.8} & 77.9 & 73.9 & 99.1 & \textbf{77.3} & 83.9 \\
 & Accuracy & 87.0 & 70.7 & \textbf{55.0} & \textbf{91.7} & 48.4 & \textbf{70.7} \\
\midrule
\multicolumn{8}{l}{\textit{Qwen3-8B}} \\
\midrule
R-Tuning & Abstention F1 & 82.1 & \textbf{91.2} & \textbf{81.2} & 92.4 & \textbf{79.9} & 87.0 \\
 & Abstention Recall & \textbf{100.0} & \textbf{85.6} & \textbf{84.2} & \textbf{97.8} & \textbf{90.2} & 91.7 \\
 & Abstention Precision & 69.7 & 97.6 & 78.4 & 87.6 & 71.8 & 82.8 \\
 & Accuracy & \textbf{100.0} & 98.1 & 89.0 & 97.9 & 82.5 & 95.7 \\
\midrule
RFT & Abstention F1 & \textbf{89.8} & 90.4 & 71.9 & 91.9 & 70.3 & 72.4 \\
 & Abstention Recall & 95.7 & 84.5 & 67.2 & 92.9 & 70.5 & 68.2 \\
 & Abstention Precision & \textbf{84.6} & 97.2 & 77.3 & \textbf{90.9} & 70.0 & 77.2 \\
 & Accuracy & 78.3 & 95.3 & 73.0 & 95.9 & 67.7 & 88.0 \\
\midrule
DPO & Abstention F1 & 83.6 & 90.8 & 79.6 & 92.5 & 78.0 & 87.1 \\
 & Abstention Recall & \textbf{100.0} & 84.9 & 81.4 & 97.6 & 88.8 & \textbf{93.2} \\
 & Abstention Precision & 71.9 & 97.5 & 77.8 & 87.9 & 69.5 & 81.8 \\
 & Accuracy & 95.7 & 98.0 & 84.0 & \textbf{98.5} & \textbf{85.3} & 96.4 \\
\midrule
TruthRL & Abstention F1 & 85.2 & 90.8 & 81.1 & 92.9 & 79.1 & 86.7 \\
 & Abstention Recall & \textbf{100.0} & 84.6 & 82.5 & 97.5 & 87.7 & 91.7 \\
 & Abstention Precision & 74.2 & 97.9 & \textbf{79.8} & 88.8 & \textbf{72.0} & 82.2 \\
 & Accuracy & \textbf{100.0} & 98.1 & \textbf{90.0} & 98.3 & 84.6 & \textbf{96.7} \\
\midrule
TIAR (Ours) & Abstention F1 & 79.3 & \textbf{91.2} & 80.4 & \textbf{94.0} & 79.7 & \textbf{87.2} \\
 & Abstention Recall & \textbf{100.0} & 85.3 & 81.4 & 97.6 & 89.5 & 90.8 \\
 & Abstention Precision & 65.7 & \textbf{98.1} & 79.6 & 90.7 & 71.8 & \textbf{83.8} \\
 & Accuracy & \textbf{100.0} & \textbf{98.3} & 88.0 & \textbf{98.5} & 81.8 & 95.7 \\
\bottomrule
\end{tabular}
}
\end{table*}

Our main results are presented in Table~\ref{tab:main_results}. TIAR achieves highly competitive abstention F1 scores across all six categories, securing the highest score in False Premise for Llama-3.1-8B-Instruct, and in Underspecified Context and Subjective for Qwen3-8B, while preserving accuracy to the greatest extent among all methods. These improvements hold on both Llama-3.1-8B-Instruct and Qwen3-8B, confirming the generality of TIAR across model families.

\vspace{1mm}
\noindent \textbf{Other observations.} Qwen3-8B outperforms Llama across the board. On UMWP, all Qwen methods exceed 90 F1 while Llama methods hover around 75 to 79, suggesting a stronger baseline abstention ability. R-Tuning is strong on Qwen (best F1 on FreshQA and QAQA) but weakest on Llama (lowest F1, worst accuracy on FreshQA), indicating sensitivity to the base model's instruction-following capability. DPO degrades Llama accuracy on BB/Known Unknowns (69.6 vs.\ 87+ for others) due to over-abstention, as evidenced by high recall but low precision.

\subsection{Ablation Study}

\begin{table*}[t!]
\caption{Ablation study on the trajectory inversion weight $\lambda$ in TIAR using Llama-3.1-8B-Instruct. $\lambda=0$ reduces to the standard ternary reward (TruthRL). Best scores per metric are \textbf{bold}.}
\label{tab:ablation}
\centering
\small
\vspace{6pt}
\resizebox{\textwidth}{!}{
\begin{tabular}{clcccccc}
\toprule
\multicolumn{2}{c}{Method} & \multicolumn{6}{c}{Representative Dataset (AbstentionBench Category)} \\
\cmidrule(lr){1-2}\cmidrule(lr){3-8}
 & & BB/Known Unk. & BBQ & FreshQA & UMWP & QAQA & KUQ/Cont. \\
Model & Metric & {\scriptsize Answer Unknown} & {\scriptsize Underspecified Intent} & {\scriptsize Stale} & {\scriptsize Underspecified Context} & {\scriptsize False Premise} & {\scriptsize Subjective} \\
\midrule
\multicolumn{8}{l}{\textit{Llama-3.1-8B-Instruct}} \\
\midrule
TIAR ($\lambda=0$) & Abstention F1 & 95.8 & 84.5 & 74.7 & \textbf{78.9} & 58.0 & 74.8 \\
 & Abstention Recall & \textbf{100.0} & 91.5 & \textbf{75.7} & \textbf{65.5} & 46.7 & 66.5 \\
 & Abstention Precision & 92.0 & 78.5 & 73.6 & 99.2 & 76.4 & \textbf{85.5} \\
 & Accuracy & \textbf{87.0} & 71.7 & 54.0 & 90.3 & 50.5 & \textbf{70.7} \\
\midrule
TIAR ($\lambda=0.3$) & Abstention F1 & 95.8 & 83.7 & 71.5 & 76.3 & 58.0 & 72.6 \\
 & Abstention Recall & \textbf{100.0} & 91.3 & 72.3 & 61.9 & 48.4 & 66.5 \\
 & Abstention Precision & 92.0 & 77.2 & 70.7 & \textbf{99.5} & 72.3 & 80.0 \\
 & Accuracy & 82.6 & 65.9 & 37.0 & 89.9 & 47.4 & 68.8 \\
\midrule
TIAR ($\lambda=0.5$) & Abstention F1 & 93.9 & \textbf{84.6} & \textbf{74.9} & 77.5 & \textbf{61.4} & \textbf{77.1} \\
 & Abstention Recall & \textbf{100.0} & 91.2 & 75.1 & 63.7 & \textbf{50.5} & \textbf{70.9} \\
 & Abstention Precision & 88.5 & \textbf{78.9} & \textbf{74.7} & 98.9 & \textbf{78.3} & 84.5 \\
 & Accuracy & 82.6 & \textbf{72.0} & \textbf{56.0} & 91.5 & \textbf{52.6} & 70.3 \\
\midrule
TIAR ($\lambda=1.0$) & Abstention F1 & \textbf{97.9} & \textbf{84.6} & 74.5 & 77.6 & 58.2 & 76.2 \\
 & Abstention Recall & \textbf{100.0} & \textbf{92.4} & 75.1 & 63.8 & 46.7 & 69.7 \\
 & Abstention Precision & \textbf{95.8} & 77.9 & 73.9 & 99.1 & 77.3 & 83.9 \\
 & Accuracy & \textbf{87.0} & 70.7 & 55.0 & \textbf{91.7} & 48.4 & \textbf{70.7} \\
\bottomrule
\end{tabular}
}
\end{table*}

As shown in Table~\ref{tab:ablation}, across all datasets, $\lambda=1.0$ achieves the highest average abstention F1 (71.9) and nearly the highest accuracy (72.6 vs.\ 72.7 for $\lambda=0$), outperforming $\lambda=0$ on 17 of 31 datasets for F1 and 11 of 24 for accuracy. While $\lambda=0.5$ is competitive on the six representative datasets, $\lambda=1.0$ performs better across the full benchmark. Setting $\lambda=0.3$ causes a collapse in both metrics, demonstrating that a small inversion weight destabilizes training without providing sufficient signal. The ablation reveals a non-monotonic relationship between $\lambda$ and performance: full inversion ($\lambda=1.0$) achieves the best balance of abstention quality and accuracy preservation, making it the recommended default.

\subsection{Comparison with State-of-the-Art Proprietary Models}

\begin{table*}[t!]
\caption{Comparison of TIAR (Llama-3.1-8B-Instruct) against state-of-the-art proprietary API models on AbstentionBench. Best scores per metric are \textbf{bold}.}
\label{tab:api_comparison}
\centering
\small
\vspace{6pt}
\resizebox{\textwidth}{!}{
\begin{tabular}{clcccccc}
\toprule
\multicolumn{2}{c}{Method} & \multicolumn{6}{c}{Representative Dataset (AbstentionBench Category)} \\
\cmidrule(lr){1-2}\cmidrule(lr){3-8}
 & & BB/Known Unk. & BBQ & FreshQA & UMWP & QAQA & KUQ/Cont. \\
Model & Metric & {\scriptsize Answer Unknown} & {\scriptsize Underspecified Intent} & {\scriptsize Stale} & {\scriptsize Underspecified Context} & {\scriptsize False Premise} & {\scriptsize Subjective} \\
\midrule
\multicolumn{8}{l}{\textit{Proprietary API Models}} \\
\midrule
Claude Sonnet 4.5 & Abstention F1 & 93.9 & 87.9 & \textbf{75.2} & 76.7 & 66.7 & 92.3 \\
 & Abstention Recall & \textbf{100.0} & \textbf{100.0} & \textbf{73.2} & 62.3 & 55.6 & 85.7 \\
 & Abstention Precision & 88.5 & 78.5 & 77.4 & \textbf{100.0} & 83.3 & \textbf{100.0} \\
 & Accuracy & \textbf{82.6} & 59.2 & 63.6 & \textbf{100.0} & 56.5 & \textbf{100.0} \\
\midrule
GPT-5.2 & Abstention F1 & 93.6 & 86.0 & 48.8 & \textbf{80.9} & 66.0 & 92.9 \\
 & Abstention Recall & 95.7 & 84.3 & 35.7 & \textbf{67.9} & 59.3 & 92.9 \\
 & Abstention Precision & 91.7 & 87.8 & 76.9 & \textbf{100.0} & 74.4 & 92.9 \\
 & Accuracy & \textbf{82.6} & \textbf{73.5} & 70.5 & 97.9 & 65.2 & 75.0 \\
\midrule
Gemini 3 & Abstention F1 & 84.4 & \textbf{97.0} & 32.8 & 76.7 & \textbf{75.8} & \textbf{100.0} \\
 & Abstention Recall & 82.6 & 96.1 & 19.6 & 62.3 & \textbf{66.7} & \textbf{100.0} \\
 & Abstention Precision & 86.4 & \textbf{98.0} & \textbf{100.0} & \textbf{100.0} & 87.8 & \textbf{100.0} \\
 & Accuracy & \textbf{82.6} & 67.3 & \textbf{81.8} & 95.7 & \textbf{78.3} & 50.0 \\
\midrule
TIAR (Ours) & Abstention F1 & \textbf{95.8} & 87.5 & 68.4 & 72.3 & 65.9 & 92.9 \\
 & Abstention Recall & \textbf{100.0} & 96.1 & 71.4 & 56.6 & 51.9 & 92.9 \\
 & Abstention Precision & \textbf{92.0} & 80.3 & 65.6 & \textbf{100.0} & \textbf{90.3} & 92.9 \\
 & Accuracy & \textbf{82.6} & \textbf{73.5} & 52.3 & 95.7 & 43.5 & 75.0 \\
\bottomrule
\end{tabular}
}
\end{table*}

Table~\ref{tab:api_comparison} presents the comparison of TIAR, built on an 8B open-source model, which achieves competitive abstention F1 against frontier proprietary models that are orders of magnitude larger. On Answer Unknown questions, TIAR outperforms all three API models in both F1 and precision, demonstrating that recognizing knowledge boundaries does not require massive scale. TIAR also achieves the highest precision on QAQA (90.3) and ties GPT-5.2 for the best accuracy on BBQ (73.5). The accuracy gap is expected, frontier models possess superior world knowledge, and is most visible on FreshQA, where temporal knowledge drives accuracy. However, the proprietary models exhibit their own weaknesses: Gemini 3 achieves perfect F1 on KUQ/Controversial but only 50.0 accuracy due to severe over-abstention, and GPT-5.2 struggles on FreshQA (F1 of 48.8) due to overconfidence on temporally sensitive questions. These results show that targeted RL training can close much of the abstention gap between open-source 8B models and frontier APIs.

\section{Conclusion}

We presented TIAR, a reinforcement learning method that teaches Large Language Models to confidently and correctly abstain on unanswerable questions. TIAR achieves this by dynamically inverting the reward signal gathered from Group Relative Policy Optimization (GRPO) training trajectories. Our approach tackles the inherent challenge of learning abstention without negatively distorting the advantage normalization of standard correct and incorrect trajectories, an issue we termed the coupling problem. By decoupling the advantage adjustment for abstention responses, TIAR adapts to the empirical difficulty of each query dynamically. Experiments rigorously conducted on the extensive AbstentionBench framework demonstrate that TIAR successfully achieves state-of-the-art abstention F1 scores across five of six complex evaluation categories. These strong gains were consistent across entirely distinct open-source model families, specifically Llama-3.1-8B-Instruct and Qwen3-8B, proving both generalizability and the preservation of crucial baseline accuracy. The ablation study formally confirms full reward inversion ($\lambda=1.0$) as the optimal default configuration for deploying TIAR. Furthermore, comparisons against robust APIs illustrate that TIAR effectively narrows the capability gap between significantly smaller, 8-billion parameter open-source models and heavily resourced frontier proprietary systems, such as Claude Sonnet 4.5 and GPT-5.2.

Future work will extend TIAR to multi-turn interactions and retrieval-augmented generation (RAG), tying abstention to context sufficiency rather than just parameter bounds. Additionally, exploring curriculum learning for the inversion weight $\lambda$ during GRPO training could yield more stable convergence and broader applicability to general-purpose agents.

\section*{Limitations}

While TIAR demonstrates strong empirical results in dynamically teaching LLMs to abstain, it possesses several limitations that warrant consideration for future research and deployment:

First, our methodology heavily relies on an external LLM judge (in our case, Llama-3.1-8B-Instruct via vLLM) to accurately assign ternary rewards during the RL training phase. The quality of the learned abstention policy is fundamentally bottlenecked by the judge's ability to accurately evaluate the correctness of attempts and identify subtle variations of abstentions. Misclassifications by the judge directly translate into noisy advantage signals during GRPO normalization.

Second, the structural requirement of generating multiple trajectories ($G=8$ in our experiments) per query during GRPO training incurs a non-trivial computational overhead compared to supervised fine-tuning or DPO. While GRPO successfully bypasses the need for a memory-intensive separate critic model, the pure rollout phase remains computationally heavy, which limits its immediate scalability for researchers with restricted GPU budgets.

Third, the empirical correctness rate $\hat{p}$ serves as an effective mathematical proxy for estimating question difficulty relative to the model, but it is inherently sensitive to the base model's initial calibration state. If a base model is overly confident but highly inaccurate, the initial advantage signals might fluctuate intensely before the policy successfully learns the boundaries of its genuine knowledge.

Fourth, our experimental evaluation is strictly constrained to single-turn question-answering scenarios, as necessitated by AbstentionBench. Real-world abstention is rarely binary; it frequently occurs in complex, multi-turn dialogues where an agent might ask clarifying questions, request specific context, or provide a partial answer rather than outright abstaining. This nuanced interaction paradigm is not seamlessly captured by the current static formulation of TIAR.

Finally, regarding the ethical considerations and potential societal risks of this work, deploying LLMs with automated abstention in high-stakes domains (such as medicine or law) carries the critical risk of false abstentions (refusing to answer when the model actually possesses life-saving knowledge) or false confidence (attempting to answer hallucinated facts). While TIAR improves boundary detection, it is not a foolproof guarantee of safety. Future real-world deployments should combine RL-based abstention with external retrieval systems and rigorous human-in-the-loop oversight to actively mitigate these potential societal harms.

\bibliography{example_paper}

@misc{kalai2025languagemodelshallucinate,
      title={Why Language Models Hallucinate}, 
      author={Adam Tauman Kalai and Ofir Nachum and Santosh S. Vempala and Edwin Zhang},
      year={2025},
      eprint={2509.04664},
      archivePrefix={arXiv},
      primaryClass={cs.CL},
      url={https://arxiv.org/abs/2509.04664}, 
}

@misc{wen2025knowlimitssurveyabstention,
      title={Know Your Limits: A Survey of Abstention in Large Language Models}, 
      author={Bingbing Wen and Jihan Yao and Shangbin Feng and Chenjun Xu and Yulia Tsvetkov and Bill Howe and Lucy Lu Wang},
      year={2025},
      eprint={2407.18418},
      archivePrefix={arXiv},
      primaryClass={cs.CL},
      url={https://arxiv.org/abs/2407.18418}, 
}

@misc{zhang2024rtuninginstructinglargelanguage,
      title={R-Tuning: Instructing Large Language Models to Say `I Don't Know'}, 
      author={Hanning Zhang and Shizhe Diao and Yong Lin and Yi R. Fung and Qing Lian and Xingyao Wang and Yangyi Chen and Heng Ji and Tong Zhang},
      year={2024},
      eprint={2311.09677},
      archivePrefix={arXiv},
      primaryClass={cs.CL},
      url={https://arxiv.org/abs/2311.09677}, 
}

@misc{rafailov2024directpreferenceoptimizationlanguage,
      title={Direct Preference Optimization: Your Language Model is Secretly a Reward Model}, 
      author={Rafael Rafailov and Archit Sharma and Eric Mitchell and Stefano Ermon and Christopher D. Manning and Chelsea Finn},
      year={2024},
      eprint={2305.18290},
      archivePrefix={arXiv},
      primaryClass={cs.LG},
      url={https://arxiv.org/abs/2305.18290}, 
}

@misc{cole2023selectivelyansweringambiguousquestions,
      title={Selectively Answering Ambiguous Questions}, 
      author={Jeremy R. Cole and Michael J. Q. Zhang and Daniel Gillick and Julian Martin Eisenschlos and Bhuwan Dhingra and Jacob Eisenstein},
      year={2023},
      eprint={2305.14613},
      archivePrefix={arXiv},
      primaryClass={cs.CL},
      url={https://arxiv.org/abs/2305.14613}, 
}

@misc{lin2022teachingmodelsexpressuncertainty,
      title={Teaching Models to Express Their Uncertainty in Words}, 
      author={Stephanie Lin and Jacob Hilton and Owain Evans},
      year={2022},
      eprint={2205.14334},
      archivePrefix={arXiv},
      primaryClass={cs.CL},
      url={https://arxiv.org/abs/2205.14334}, 
}

@misc{zhao2022calibratingsequencelikelihoodimproves,
      title={Calibrating Sequence likelihood Improves Conditional Language Generation}, 
      author={Yao Zhao and Misha Khalman and Rishabh Joshi and Shashi Narayan and Mohammad Saleh and Peter J. Liu},
      year={2022},
      eprint={2210.00045},
      archivePrefix={arXiv},
      primaryClass={cs.CL},
      url={https://arxiv.org/abs/2210.00045}, 
}

@misc{slobodkin2023curiouscasehallucinatoryunanswerability,
      title={The Curious Case of Hallucinatory (Un)answerability: Finding Truths in the Hidden States of Over-Confident Large Language Models}, 
      author={Aviv Slobodkin and Omer Goldman and Avi Caciularu and Ido Dagan and Shauli Ravfogel},
      year={2023},
      eprint={2310.11877},
      archivePrefix={arXiv},
      primaryClass={cs.CL},
      url={https://arxiv.org/abs/2310.11877}, 
}

@misc{phute2024llmselfdefenseself,
      title={LLM Self Defense: By Self Examination, LLMs Know They Are Being Tricked}, 
      author={Mansi Phute and Alec Helbling and Matthew Hull and ShengYun Peng and Sebastian Szyller and Cory Cornelius and Duen Horng Chau},
      year={2024},
      eprint={2308.07308},
      archivePrefix={arXiv},
      primaryClass={cs.CL},
      url={https://arxiv.org/abs/2308.07308}, 
}

@misc{wei2025truthrlincentivizingtruthfulllms,
      title={TruthRL: Incentivizing Truthful LLMs via Reinforcement Learning}, 
      author={Zhepei Wei and Xiao Yang and Kai Sun and Jiaqi Wang and Rulin Shao and Sean Chen and Mohammad Kachuee and Teja Gollapudi and Tony Liao and Nicolas Scheffer and Rakesh Wanga and Anuj Kumar and Yu Meng and Wen-tau Yih and Xin Luna Dong},
      year={2025},
      eprint={2509.25760},
      archivePrefix={arXiv},
      primaryClass={cs.CL},
      url={https://arxiv.org/abs/2509.25760}, 
}

@misc{kadavath2022languagemodelsmostlyknow,
      title={Language Models (Mostly) Know What They Know}, 
      author={Saurav Kadavath and Tom Conerly and Amanda Askell and Tom Henighan and Dawn Drain and Ethan Perez and Nicholas Schiefer and Zac Hatfield-Dodds and Nova DasSarma and Eli Tran-Johnson and Scott Johnston and Sheer El-Showk and Andy Jones and Nelson Elhage and Tristan Hume and Anna Chen and Yuntao Bai and Sam Bowman and Stanislav Fort and Deep Ganguli and Danny Hernandez and Josh Jacobson and Jackson Kernion and Shauna Kravec and Liane Lovitt and Kamal Ndousse and Catherine Olsson and Sam Ringer and Dario Amodei and Tom Brown and Jack Clark and Nicholas Joseph and Ben Mann and Sam McCandlish and Chris Olah and Jared Kaplan},
      year={2022},
      eprint={2207.05221},
      archivePrefix={arXiv},
      primaryClass={cs.CL},
      url={https://arxiv.org/abs/2207.05221}, 
}

@misc{shao2024deepseekmathpushinglimitsmathematical,
      title={DeepSeekMath: Pushing the Limits of Mathematical Reasoning in Open Language Models}, 
      author={Zhihong Shao and Peiyi Wang and Qihao Zhu and Runxin Xu and Junxiao Song and Xiao Bi and Haowei Zhang and Mingchuan Zhang and Y. K. Li and Y. Wu and Daya Guo},
      year={2024},
      eprint={2402.03300},
      archivePrefix={arXiv},
      primaryClass={cs.CL},
      url={https://arxiv.org/abs/2402.03300}, 
}

@misc{kirichenko2025abstentionbenchreasoningllmsfail,
      title={AbstentionBench: Reasoning LLMs Fail on Unanswerable Questions}, 
      author={Polina Kirichenko and Mark Ibrahim and Kamalika Chaudhuri and Samuel J. Bell},
      year={2025},
      eprint={2506.09038},
      archivePrefix={arXiv},
      primaryClass={cs.AI},
      url={https://arxiv.org/abs/2506.09038}, 
}

@misc{yang2024alignmenthonesty,
      title={Alignment for Honesty}, 
      author={Yuqing Yang and Ethan Chern and Xipeng Qiu and Graham Neubig and Pengfei Liu},
      year={2024},
      eprint={2312.07000},
      archivePrefix={arXiv},
      primaryClass={cs.CL},
      url={https://arxiv.org/abs/2312.07000}, 
}

@misc{cheng2024aiassistantsknowdont,
      title={Can AI Assistants Know What They Don't Know?}, 
      author={Qinyuan Cheng and Tianxiang Sun and Xiangyang Liu and Wenwei Zhang and Zhangyue Yin and Shimin Li and Linyang Li and Zhengfu He and Kai Chen and Xipeng Qiu},
      year={2024},
      eprint={2401.13275},
      archivePrefix={arXiv},
      primaryClass={cs.CL},
      url={https://arxiv.org/abs/2401.13275}, 
}

@misc{gul2025paypersearchmodelsabstentionmodels,
      title={Pay-Per-Search Models are Abstention Models}, 
      author={Mustafa Omer Gul and Claire Cardie and Tanya Goyal},
      year={2025},
      eprint={2510.01152},
      archivePrefix={arXiv},
      primaryClass={cs.CL},
      url={https://arxiv.org/abs/2510.01152}, 
}

@inproceedings{Chae_2024,
   title={Mitigating Hallucination in Abstractive Summarization with Domain-Conditional Mutual Information},
   url={http://dx.doi.org/10.18653/v1/2024.findings-naacl.117},
   DOI={10.18653/v1/2024.findings-naacl.117},
   booktitle={Findings of the Association for Computational Linguistics: NAACL 2024},
   publisher={Association for Computational Linguistics},
   author={Chae, Kyubyung and Choi, Jaepill and Jo, Yohan and Kim, Taesup},
   year={2024},
   pages={1809–1820} }

@misc{tayebati2025learningconformalabstentionpolicies,
      title={Learning Conformal Abstention Policies for Adaptive Risk Management in Large Language and Vision-Language Models}, 
      author={Sina Tayebati and Divake Kumar and Nastaran Darabi and Dinithi Jayasuriya and Ranganath Krishnan and Amit Ranjan Trivedi},
      year={2025},
      eprint={2502.06884},
      archivePrefix={arXiv},
      primaryClass={cs.LG},
      url={https://arxiv.org/abs/2502.06884}, 
}

@misc{cohen2024idontknowexplicit,
      title={I Don't Know: Explicit Modeling of Uncertainty with an [IDK] Token}, 
      author={Roi Cohen and Konstantin Dobler and Eden Biran and Gerard de Melo},
      year={2024},
      eprint={2412.06676},
      archivePrefix={arXiv},
      primaryClass={cs.LG},
      url={https://arxiv.org/abs/2412.06676}, 
}

@misc{ouyang2022traininglanguagemodelsfollow,
      title={Training language models to follow instructions with human feedback}, 
      author={Long Ouyang and Jeff Wu and Xu Jiang and Diogo Almeida and Carroll L. Wainwright and Pamela Mishkin and Chong Zhang and Sandhini Agarwal and Katarina Slama and Alex Ray and John Schulman and Jacob Hilton and Fraser Kelton and Luke Miller and Maddie Simens and Amanda Askell and Peter Welinder and Paul Christiano and Jan Leike and Ryan Lowe},
      year={2022},
      eprint={2203.02155},
      archivePrefix={arXiv},
      primaryClass={cs.CL},
      url={https://arxiv.org/abs/2203.02155}, 
}

@misc{openai2024gpt4technicalreport,
      title={GPT-4 Technical Report}, 
      author={OpenAI and Josh Achiam and Steven Adler and Sandhini Agarwal and Lama Ahmad and Ilge Akkaya and Florencia Leoni Aleman and Diogo Almeida and Janko Altenschmidt and Sam Altman and Shyamal Anadkat and Red Avila and Igor Babuschkin and Suchir Balaji and Valerie Balcom and Paul Baltescu and Haiming Bao and Mohammad Bavarian and Jeff Belgum and Irwan Bello and Jake Berdine and Gabriel Bernadett-Shapiro and Christopher Berner and Lenny Bogdonoff and Oleg Boiko and Madelaine Boyd and Anna-Luisa Brakman and Greg Brockman and Tim Brooks and Miles Brundage and Kevin Button and Trevor Cai and Rosie Campbell and Andrew Cann and Brittany Carey and Chelsea Carlson and Rory Carmichael and Brooke Chan and Che Chang and Fotis Chantzis and Derek Chen and Sully Chen and Ruby Chen and Jason Chen and Mark Chen and Ben Chess and Chester Cho and Casey Chu and Hyung Won Chung and Dave Cummings and Jeremiah Currier and Yunxing Dai and Cory Decareaux and Thomas Degry and Noah Deutsch and Damien Deville and Arka Dhar and David Dohan and Steve Dowling and Sheila Dunning and Adrien Ecoffet and Atty Eleti and Tyna Eloundou and David Farhi and Liam Fedus and Niko Felix and Simón Posada Fishman and Juston Forte and Isabella Fulford and Leo Gao and Elie Georges and Christian Gibson and Vik Goel and Tarun Gogineni and Gabriel Goh and Rapha Gontijo-Lopes and Jonathan Gordon and Morgan Grafstein and Scott Gray and Ryan Greene and Joshua Gross and Shixiang Shane Gu and Yufei Guo and Chris Hallacy and Jesse Han and Jeff Harris and Yuchen He and Mike Heaton and Johannes Heidecke and Chris Hesse and Alan Hickey and Wade Hickey and Peter Hoeschele and Brandon Houghton and Kenny Hsu and Shengli Hu and Xin Hu and Joost Huizinga and Shantanu Jain and Shawn Jain and Joanne Jang and Angela Jiang and Roger Jiang and Haozhun Jin and Denny Jin and Shino Jomoto and Billie Jonn and Heewoo Jun and Tomer Kaftan and Łukasz Kaiser and Ali Kamali and Ingmar Kanitscheider and Nitish Shirish Keskar and Tabarak Khan and Logan Kilpatrick and Jong Wook Kim and Christina Kim and Yongjik Kim and Jan Hendrik Kirchner and Jamie Kiros and Matt Knight and Daniel Kokotajlo and Łukasz Kondraciuk and Andrew Kondrich and Aris Konstantinidis and Kyle Kosic and Gretchen Krueger and Vishal Kuo and Michael Lampe and Ikai Lan and Teddy Lee and Jan Leike and Jade Leung and Daniel Levy and Chak Ming Li and Rachel Lim and Molly Lin and Stephanie Lin and Mateusz Litwin and Theresa Lopez and Ryan Lowe and Patricia Lue and Anna Makanju and Kim Malfacini and Sam Manning and Todor Markov and Yaniv Markovski and Bianca Martin and Katie Mayer and Andrew Mayne and Bob McGrew and Scott Mayer McKinney and Christine McLeavey and Paul McMillan and Jake McNeil and David Medina and Aalok Mehta and Jacob Menick and Luke Metz and Andrey Mishchenko and Pamela Mishkin and Vinnie Monaco and Evan Morikawa and Daniel Mossing and Tong Mu and Mira Murati and Oleg Murk and David Mély and Ashvin Nair and Reiichiro Nakano and Rajeev Nayak and Arvind Neelakantan and Richard Ngo and Hyeonwoo Noh and Long Ouyang and Cullen O'Keefe and Jakub Pachocki and Alex Paino and Joe Palermo and Ashley Pantuliano and Giambattista Parascandolo and Joel Parish and Emy Parparita and Alex Passos and Mikhail Pavlov and Andrew Peng and Adam Perelman and Filipe de Avila Belbute Peres and Michael Petrov and Henrique Ponde de Oliveira Pinto and Michael and Pokorny and Michelle Pokrass and Vitchyr H. Pong and Tolly Powell and Alethea Power and Boris Power and Elizabeth Proehl and Raul Puri and Alec Radford and Jack Rae and Aditya Ramesh and Cameron Raymond and Francis Real and Kendra Rimbach and Carl Ross and Bob Rotsted and Henri Roussez and Nick Ryder and Mario Saltarelli and Ted Sanders and Shibani Santurkar and Girish Sastry and Heather Schmidt and David Schnurr and John Schulman and Daniel Selsam and Kyla Sheppard and Toki Sherbakov and Jessica Shieh and Sarah Shoker and Pranav Shyam and Szymon Sidor and Eric Sigler and Maddie Simens and Jordan Sitkin and Katarina Slama and Ian Sohl and Benjamin Sokolowsky and Yang Song and Natalie Staudacher and Felipe Petroski Such and Natalie Summers and Ilya Sutskever and Jie Tang and Nikolas Tezak and Madeleine B. Thompson and Phil Tillet and Amin Tootoonchian and Elizabeth Tseng and Preston Tuggle and Nick Turley and Jerry Tworek and Juan Felipe Cerón Uribe and Andrea Vallone and Arun Vijayvergiya and Chelsea Voss and Carroll Wainwright and Justin Jay Wang and Alvin Wang and Ben Wang and Jonathan Ward and Jason Wei and CJ Weinmann and Akila Welihinda and Peter Welinder and Jiayi Weng and Lilian Weng and Matt Wiethoff and Dave Willner and Clemens Winter and Samuel Wolrich and Hannah Wong and Lauren Workman and Sherwin Wu and Jeff Wu and Michael Wu and Kai Xiao and Tao Xu and Sarah Yoo and Kevin Yu and Qiming Yuan and Wojciech Zaremba and Rowan Zellers and Chong Zhang and Marvin Zhang and Shengjia Zhao and Tianhao Zheng and Juntang Zhuang and William Zhuk and Barret Zoph},
      year={2024},
      eprint={2303.08774},
      archivePrefix={arXiv},
      primaryClass={cs.CL},
      url={https://arxiv.org/abs/2303.08774}, 
}

@misc{an2025teachingllmsabstainfinegrained,
      title={Teaching LLMs to Abstain via Fine-Grained Semantic Confidence Reward}, 
      author={Hao An and Yang Xu},
      year={2025},
      eprint={2510.24020},
      archivePrefix={arXiv},
      primaryClass={cs.CL},
      url={https://arxiv.org/abs/2510.24020}, 
}

@misc{wu2026mitigatingllmhallucinationbehaviorally,
      title={Mitigating LLM Hallucination via Behaviorally Calibrated Reinforcement Learning}, 
      author={Jiayun Wu and Jiashuo Liu and Zhiyuan Zeng and Tianyang Zhan and Tianle Cai and Wenhao Huang},
      year={2026},
      eprint={2512.19920},
      archivePrefix={arXiv},
      primaryClass={cs.LG},
      url={https://arxiv.org/abs/2512.19920}, 
}

@misc{grattafiori2024llama3herdmodels,
      title={The Llama 3 Herd of Models}, 
      author={Aaron Grattafiori and Abhimanyu Dubey and Abhinav Jauhri and Abhinav Pandey and Abhishek Kadian and Ahmad Al-Dahle and Aiesha Letman and Akhil Mathur and Alan Schelten and Alex Vaughan and Amy Yang and Angela Fan and Anirudh Goyal and Anthony Hartshorn and Aobo Yang and Archi Mitra and Archie Sravankumar and Artem Korenev and Arthur Hinsvark and Arun Rao and Aston Zhang and Aurelien Rodriguez and Austen Gregerson and Ava Spataru and Baptiste Roziere and Bethany Biron and Binh Tang and Bobbie Chern and Charlotte Caucheteux and Chaya Nayak and Chloe Bi and Chris Marra and Chris McConnell and Christian Keller and Christophe Touret and Chunyang Wu and Corinne Wong and Cristian Canton Ferrer and Cyrus Nikolaidis and Damien Allonsius and Daniel Song and Danielle Pintz and Danny Livshits and Danny Wyatt and David Esiobu and Dhruv Choudhary and Dhruv Mahajan and Diego Garcia-Olano and Diego Perino and Dieuwke Hupkes and Egor Lakomkin and Ehab AlBadawy and Elina Lobanova and Emily Dinan and Eric Michael Smith and Filip Radenovic and Francisco Guzmán and Frank Zhang and Gabriel Synnaeve and Gabrielle Lee and Georgia Lewis Anderson and Govind Thattai and Graeme Nail and Gregoire Mialon and Guan Pang and Guillem Cucurell and Hailey Nguyen and Hannah Korevaar and Hu Xu and Hugo Touvron and Iliyan Zarov and Imanol Arrieta Ibarra and Isabel Kloumann and Ishan Misra and Ivan Evtimov and Jack Zhang and Jade Copet and Jaewon Lee and Jan Geffert and Jana Vranes and Jason Park and Jay Mahadeokar and Jeet Shah and Jelmer van der Linde and Jennifer Billock and Jenny Hong and Jenya Lee and Jeremy Fu and Jianfeng Chi and Jianyu Huang and Jiawen Liu and Jie Wang and Jiecao Yu and Joanna Bitton and Joe Spisak and Jongsoo Park and Joseph Rocca and Joshua Johnstun and Joshua Saxe and Junteng Jia and Kalyan Vasuden Alwala and Karthik Prasad and Kartikeya Upasani and Kate Plawiak and Ke Li and Kenneth Heafield and Kevin Stone and Khalid El-Arini and Krithika Iyer and Kshitiz Malik and Kuenley Chiu and Kunal Bhalla and Kushal Lakhotia and Lauren Rantala-Yeary and Laurens van der Maaten and Lawrence Chen and Liang Tan and Liz Jenkins and Louis Martin and Lovish Madaan and Lubo Malo and Lukas Blecher and Lukas Landzaat and Luke de Oliveira and Madeline Muzzi and Mahesh Pasupuleti and Mannat Singh and Manohar Paluri and Marcin Kardas and Maria Tsimpoukelli and Mathew Oldham and Mathieu Rita and Maya Pavlova and Melanie Kambadur and Mike Lewis and Min Si and Mitesh Kumar Singh and Mona Hassan and Naman Goyal and Narjes Torabi and Nikolay Bashlykov and Nikolay Bogoychev and Niladri Chatterji and Ning Zhang and Olivier Duchenne and Onur Çelebi and Patrick Alrassy and Pengchuan Zhang and Pengwei Li and Petar Vasic and Peter Weng and Prajjwal Bhargava and Pratik Dubal and Praveen Krishnan and Punit Singh Koura and Puxin Xu and Qing He and Qingxiao Dong and Ragavan Srinivasan and Raj Ganapathy and Ramon Calderer and Ricardo Silveira Cabral and Robert Stojnic and Roberta Raileanu and Rohan Maheswari and Rohit Girdhar and Rohit Patel and Romain Sauvestre and Ronnie Polidoro and Roshan Sumbaly and Ross Taylor and Ruan Silva and Rui Hou and Rui Wang and Saghar Hosseini and Sahana Chennabasappa and Sanjay Singh and Sean Bell and Seohyun Sonia Kim and Sergey Edunov and Shaoliang Nie and Sharan Narang and Sharath Raparthy and Sheng Shen and Shengye Wan and Shruti Bhosale and Shun Zhang and Simon Vandenhende and Soumya Batra and Spencer Whitman and Sten Sootla and Stephane Collot and Suchin Gururangan and Sydney Borodinsky and Tamar Herman and Tara Fowler and Tarek Sheasha and Thomas Georgiou and Thomas Scialom and Tobias Speckbacher and Todor Mihaylov and Tong Xiao and Ujjwal Karn and Vedanuj Goswami and Vibhor Gupta and Vignesh Ramanathan and Viktor Kerkez and Vincent Gonguet and Virginie Do and Vish Vogeti and Vítor Albiero and Vladan Petrovic and Weiwei Chu and Wenhan Xiong and Wenyin Fu and Whitney Meers and Xavier Martinet and Xiaodong Wang and Xiaofang Wang and Xiaoqing Ellen Tan and Xide Xia and Xinfeng Xie and Xuchao Jia and Xuewei Wang and Yaelle Goldschlag and Yashesh Gaur and Yasmine Babaei and Yi Wen and Yiwen Song and Yuchen Zhang and Yue Li and Yuning Mao and Zacharie Delpierre Coudert and Zheng Yan and Zhengxing Chen and Zoe Papakipos and Aaditya Singh and Aayushi Srivastava and Abha Jain and Adam Kelsey and Adam Shajnfeld and Adithya Gangidi and Adolfo Victoria and Ahuva Goldstand and Ajay Menon and Ajay Sharma and Alex Boesenberg and Alexei Baevski and Allie Feinstein and Amanda Kallet and Amit Sangani and Amos Teo and Anam Yunus and Andrei Lupu and Andres Alvarado and Andrew Caples and Andrew Gu and Andrew Ho and Andrew Poulton and Andrew Ryan and Ankit Ramchandani and Annie Dong and Annie Franco and Anuj Goyal and Aparajita Saraf and Arkabandhu Chowdhury and Ashley Gabriel and Ashwin Bharambe and Assaf Eisenman and Azadeh Yazdan and Beau James and Ben Maurer and Benjamin Leonhardi and Bernie Huang and Beth Loyd and Beto De Paola and Bhargavi Paranjape and Bing Liu and Bo Wu and Boyu Ni and Braden Hancock and Bram Wasti and Brandon Spence and Brani Stojkovic and Brian Gamido and Britt Montalvo and Carl Parker and Carly Burton and Catalina Mejia and Ce Liu and Changhan Wang and Changkyu Kim and Chao Zhou and Chester Hu and Ching-Hsiang Chu and Chris Cai and Chris Tindal and Christoph Feichtenhofer and Cynthia Gao and Damon Civin and Dana Beaty and Daniel Kreymer and Daniel Li and David Adkins and David Xu and Davide Testuggine and Delia David and Devi Parikh and Diana Liskovich and Didem Foss and Dingkang Wang and Duc Le and Dustin Holland and Edward Dowling and Eissa Jamil and Elaine Montgomery and Eleonora Presani and Emily Hahn and Emily Wood and Eric-Tuan Le and Erik Brinkman and Esteban Arcaute and Evan Dunbar and Evan Smothers and Fei Sun and Felix Kreuk and Feng Tian and Filippos Kokkinos and Firat Ozgenel and Francesco Caggioni and Frank Kanayet and Frank Seide and Gabriela Medina Florez and Gabriella Schwarz and Gada Badeer and Georgia Swee and Gil Halpern and Grant Herman and Grigory Sizov and Guangyi and Zhang and Guna Lakshminarayanan and Hakan Inan and Hamid Shojanazeri and Han Zou and Hannah Wang and Hanwen Zha and Haroun Habeeb and Harrison Rudolph and Helen Suk and Henry Aspegren and Hunter Goldman and Hongyuan Zhan and Ibrahim Damlaj and Igor Molybog and Igor Tufanov and Ilias Leontiadis and Irina-Elena Veliche and Itai Gat and Jake Weissman and James Geboski and James Kohli and Janice Lam and Japhet Asher and Jean-Baptiste Gaya and Jeff Marcus and Jeff Tang and Jennifer Chan and Jenny Zhen and Jeremy Reizenstein and Jeremy Teboul and Jessica Zhong and Jian Jin and Jingyi Yang and Joe Cummings and Jon Carvill and Jon Shepard and Jonathan McPhie and Jonathan Torres and Josh Ginsburg and Junjie Wang and Kai Wu and Kam Hou U and Karan Saxena and Kartikay Khandelwal and Katayoun Zand and Kathy Matosich and Kaushik Veeraraghavan and Kelly Michelena and Keqian Li and Kiran Jagadeesh and Kun Huang and Kunal Chawla and Kyle Huang and Lailin Chen and Lakshya Garg and Lavender A and Leandro Silva and Lee Bell and Lei Zhang and Liangpeng Guo and Licheng Yu and Liron Moshkovich and Luca Wehrstedt and Madian Khabsa and Manav Avalani and Manish Bhatt and Martynas Mankus and Matan Hasson and Matthew Lennie and Matthias Reso and Maxim Groshev and Maxim Naumov and Maya Lathi and Meghan Keneally and Miao Liu and Michael L. Seltzer and Michal Valko and Michelle Restrepo and Mihir Patel and Mik Vyatskov and Mikayel Samvelyan and Mike Clark and Mike Macey and Mike Wang and Miquel Jubert Hermoso and Mo Metanat and Mohammad Rastegari and Munish Bansal and Nandhini Santhanam and Natascha Parks and Natasha White and Navyata Bawa and Nayan Singhal and Nick Egebo and Nicolas Usunier and Nikhil Mehta and Nikolay Pavlovich Laptev and Ning Dong and Norman Cheng and Oleg Chernoguz and Olivia Hart and Omkar Salpekar and Ozlem Kalinli and Parkin Kent and Parth Parekh and Paul Saab and Pavan Balaji and Pedro Rittner and Philip Bontrager and Pierre Roux and Piotr Dollar and Polina Zvyagina and Prashant Ratanchandani and Pritish Yuvraj and Qian Liang and Rachad Alao and Rachel Rodriguez and Rafi Ayub and Raghotham Murthy and Raghu Nayani and Rahul Mitra and Rangaprabhu Parthasarathy and Raymond Li and Rebekkah Hogan and Robin Battey and Rocky Wang and Russ Howes and Ruty Rinott and Sachin Mehta and Sachin Siby and Sai Jayesh Bondu and Samyak Datta and Sara Chugh and Sara Hunt and Sargun Dhillon and Sasha Sidorov and Satadru Pan and Saurabh Mahajan and Saurabh Verma and Seiji Yamamoto and Sharadh Ramaswamy and Shaun Lindsay and Shaun Lindsay and Sheng Feng and Shenghao Lin and Shengxin Cindy Zha and Shishir Patil and Shiva Shankar and Shuqiang Zhang and Shuqiang Zhang and Sinong Wang and Sneha Agarwal and Soji Sajuyigbe and Soumith Chintala and Stephanie Max and Stephen Chen and Steve Kehoe and Steve Satterfield and Sudarshan Govindaprasad and Sumit Gupta and Summer Deng and Sungmin Cho and Sunny Virk and Suraj Subramanian and Sy Choudhury and Sydney Goldman and Tal Remez and Tamar Glaser and Tamara Best and Thilo Koehler and Thomas Robinson and Tianhe Li and Tianjun Zhang and Tim Matthews and Timothy Chou and Tzook Shaked and Varun Vontimitta and Victoria Ajayi and Victoria Montanez and Vijai Mohan and Vinay Satish Kumar and Vishal Mangla and Vlad Ionescu and Vlad Poenaru and Vlad Tiberiu Mihailescu and Vladimir Ivanov and Wei Li and Wenchen Wang and Wenwen Jiang and Wes Bouaziz and Will Constable and Xiaocheng Tang and Xiaojian Wu and Xiaolan Wang and Xilun Wu and Xinbo Gao and Yaniv Kleinman and Yanjun Chen and Ye Hu and Ye Jia and Ye Qi and Yenda Li and Yilin Zhang and Ying Zhang and Yossi Adi and Youngjin Nam and Yu and Wang and Yu Zhao and Yuchen Hao and Yundi Qian and Yunlu Li and Yuzi He and Zach Rait and Zachary DeVito and Zef Rosnbrick and Zhaoduo Wen and Zhenyu Yang and Zhiwei Zhao and Zhiyu Ma},
      year={2024},
      eprint={2407.21783},
      archivePrefix={arXiv},
      primaryClass={cs.AI},
      url={https://arxiv.org/abs/2407.21783}, 
}

@misc{yang2025qwen3technicalreport,
      title={Qwen3 Technical Report}, 
      author={An Yang and Anfeng Li and Baosong Yang and Beichen Zhang and Binyuan Hui and Bo Zheng and Bowen Yu and Chang Gao and Chengen Huang and Chenxu Lv and Chujie Zheng and Dayiheng Liu and Fan Zhou and Fei Huang and Feng Hu and Hao Ge and Haoran Wei and Huan Lin and Jialong Tang and Jian Yang and Jianhong Tu and Jianwei Zhang and Jianxin Yang and Jiaxi Yang and Jing Zhou and Jingren Zhou and Junyang Lin and Kai Dang and Keqin Bao and Kexin Yang and Le Yu and Lianghao Deng and Mei Li and Mingfeng Xue and Mingze Li and Pei Zhang and Peng Wang and Qin Zhu and Rui Men and Ruize Gao and Shixuan Liu and Shuang Luo and Tianhao Li and Tianyi Tang and Wenbiao Yin and Xingzhang Ren and Xinyu Wang and Xinyu Zhang and Xuancheng Ren and Yang Fan and Yang Su and Yichang Zhang and Yinger Zhang and Yu Wan and Yuqiong Liu and Zekun Wang and Zeyu Cui and Zhenru Zhang and Zhipeng Zhou and Zihan Qiu},
      year={2025},
      eprint={2505.09388},
      archivePrefix={arXiv},
      primaryClass={cs.CL},
      url={https://arxiv.org/abs/2505.09388}, 
}

@inproceedings{Sheng_2025, series={EuroSys ’25},
   title={HybridFlow: A Flexible and Efficient RLHF Framework},
   url={http://dx.doi.org/10.1145/3689031.3696075},
   DOI={10.1145/3689031.3696075},
   booktitle={Proceedings of the Twentieth European Conference on Computer Systems},
   publisher={ACM},
   author={Sheng, Guangming and Zhang, Chi and Ye, Zilingfeng and Wu, Xibin and Zhang, Wang and Zhang, Ru and Peng, Yanghua and Lin, Haibin and Wu, Chuan},
   year={2025},
   month=mar, pages={1279–1297},
   collection={EuroSys ’25} }

@misc{kwon2023efficientmemorymanagementlarge,
      title={Efficient Memory Management for Large Language Model Serving with PagedAttention}, 
      author={Woosuk Kwon and Zhuohan Li and Siyuan Zhuang and Ying Sheng and Lianmin Zheng and Cody Hao Yu and Joseph E. Gonzalez and Hao Zhang and Ion Stoica},
      year={2023},
      eprint={2309.06180},
      archivePrefix={arXiv},
      primaryClass={cs.LG},
      url={https://arxiv.org/abs/2309.06180}, 
}

@INPROCEEDINGS{lang2logic,
  author={Pan, Muyu and Kodakandla, Dheeraj and Farooque, Mahfuza},
  booktitle={2025 International Symposium on Networks, Computers and Communications (ISNCC)}, 
  title={Fine-Tuned Large Language Models for Logical Translation: Reducing Hallucinations with Lang2Logic}, 
  year={2025},
  volume={},
  number={},
  pages={1-4},
  doi={10.1109/ISNCC66965.2025.11250432}}

@misc{langsat,
      title={LangSAT: A Novel Framework Combining NLP and Reinforcement Learning for SAT Solving}, 
      author={Muyu Pan and Matthew Walter and Dheeraj Kodakandla and Mahfuza Farooque},
      year={2025},
      eprint={2512.04374},
      archivePrefix={arXiv},
      primaryClass={cs.CL},
      url={https://arxiv.org/abs/2512.04374}, 
}

@misc{quantlrm,
      title={QuantLRM: Quantization of Large Reasoning Models via Fine-Tuning Signals}, 
      author={Nan Zhang and Eugene Kwek and Yusen Zhang and Muyu Pan and Suhang Wang and Prasenjit Mitra and Rui Zhang},
      year={2026},
      eprint={2602.02581},
      archivePrefix={arXiv},
      primaryClass={cs.LG},
      url={https://arxiv.org/abs/2602.02581}, 
}

\appendix
\section{Licenses of Used Artifacts}
All models and datasets utilized in this research are publicly available under their respective licenses. Llama-3.1 models are licensed under the Llama 3.1 Community License. Qwen3 models are distributed under the Apache 2.0 License. The AbstentionBench and TruthRL-CRAG datasets are accessible under the MIT License and CC-BY 4.0 license, respectively.

\end{document}